\documentclass{article}
\usepackage{arxiv}
\usepackage{graphicx}
\usepackage{wrapfig}

\usepackage{blindtext} 

\usepackage[sc]{mathpazo} 
\usepackage[T1]{fontenc} 
\usepackage{microtype} 

\usepackage[english]{babel} 

\usepackage[hang, small,labelfont=bf,up,textfont=it,up]{caption} 
\usepackage{booktabs} 

\usepackage{lettrine} 

\usepackage{titlesec}
\usepackage{enumitem} 
\setlist[itemize]{noitemsep} 

\usepackage{abstract} 

\usepackage{fancyhdr} 
\usepackage{titling} 
\usepackage{authblk}
\usepackage{verbatim}

\usepackage{hyperref} 

\usepackage{graphicx}
\usepackage{subcaption}
\usepackage{amsmath}

\pretitle{\begin{center}\Large\bfseries} 
\posttitle{\end{center}} 
\title{Recursive Decoding: A Situated Cognition Approach to Compositional Generation in Grounded Language Understanding} 
\date{}

\author{Matthew Setzler\thanks{corresponding author: matt.setzler@pnnl.gov}}
\author{Scott Howland}
\author{Lauren Phillips}
\affil{Pacific Northwest National Laboratory}

\begin{document}

\maketitle


\begin{abstract}
Compositional generalization is a troubling blind spot for neural language models. Recent efforts have presented techniques for improving a model's ability to encode novel combinations of known inputs, but less work has focused on generating novel combinations of known outputs. Here we focus on this latter ``decode-side'' form of generalization in the context of gSCAN, a synthetic benchmark for compositional generalization in grounded language understanding. We present Recursive Decoding (RD), a novel procedure for training and using seq2seq models, targeted towards decode-side generalization. Rather than generating an entire output sequence in one pass, models are trained to predict one token at a time. Inputs (i.e., the external gSCAN environment) are then incrementally updated based on predicted tokens, and re-encoded for the next decoder time step. RD thus decomposes a complex, out-of-distribution sequence generation task into a series of incremental predictions that each resemble what the model has already seen during training. RD yields dramatic improvement on two previously neglected generalization tasks in gSCAN. We provide analyses to elucidate these gains over failure of a baseline, and then discuss implications for generalization in naturalistic grounded language understanding, and seq2seq more generally.
\end{abstract}

\section{Introduction}
\label{sec:introduction}

For many years, \textit{compositional generalization} -- the ability to systematically combine known linguistic elements in novel ways -- has been a ``holy grail'' in NLP research. This ability comes quite naturally to human language learners, and thus appears to be a central inductive bias of the human learning system~\cite{chomsky2009syntactic, lake2014towards, lake2019human}. However, a range of curated datasets have revealed that standard deep neural networks struggle to generalize compositionally to novel utterances not seen during training~\cite{shridhar2020alfred, kim2020cogs, lake2018generalization}. 

Despite a proliferation of research investigating compositional generalization in deep networks, relatively little attention has been paid to compositional generalization in \textit{grounded language understanding} tasks. Grounding in language is an acknowledgement that in the real-world, language is most often used in the context of a shared physical environment. Understanding a phrase like ``hand me the big apple on the yellow table'' requires an understanding of the visual scene in which it is uttered (i.e., where is the yellow table? which apple is big?). Grounding is known to play an important role in human language acquisition and may favor compositional representations~\cite{smith2005development}.

One dataset which highlights the combination of grounding and composition is gSCAN (grounded SCAN)~\cite{ruis2020benchmark}, which extends the popular synthetic sequence-to-sequence (seq2seq) benchmark, SCAN~\cite{lake2018generalization} to include a grid-world environment. The model is presented with an input command (e.g., ``walk to the red circle'') and a grid world comprising an agent and various objects (e.g., red circle, green circle) with the goal of producing an output action sequence that completes the input command. Like SCAN, gSCAN comprises a number of carefully curated generalization tasks which have proven difficult for baseline model architectures~\cite{ruis2020benchmark}. 

Approaches to gSCAN have predominantly focused on dealing with novel combinations of known inputs – either through encoding the grid world as a network, composing specialized modules based in the input command, or decoupling the subtasks of identifying a target object from that of producing an appropriate action sequence \cite{gao2020systematic, kuo2020compositional, heinze2020think}. These innovations have boosted performance on a subset of tasks targeted towards encoding novel combinations of known inputs, but have offered virtually no improvement on the remaining tasks, underscoring the notion that different techniques may be required to address different \textit{kinds} of composition~\cite{kamp1995prototype, dankers2021paradox}. In this paper we focus on two particular tasks that have not been improved by existing efforts: ``novel direction" and ``length extrapolation". In contrast to other gSCAN tasks that previous efforts successfully improved upon, these two tasks require models to generate output sequences that are out of distribution from those produced during training, a challenge we refer to as \textit{decode-side generalization}.

To address how seq2seq models can generalize compositionally not at the input level, but based on the output sequence, we draw inspiration from the embodied/situated cognition paradigm in cognitive science~\cite{calvo2008handbook}. This paradigm offers that cognition is seldom confined to individual brains working in isolation. Rather, humans offload cognitive processing onto the external world; we draw diagrams and manipulate symbolic expressions on whiteboards to solve abstract problems, record numbers and figures on paper so we don't have to remember them, and use our hands to perform basic arithmetic \cite{chiel1997brain, clark1998extended, marghetis2016mastering}. In so doing, we recruit the external world in our thinking process, which alleviates working memory and representational demands, and allows us to incrementally reason through a problem with the help of stable (but manipulable) representations

Inspired by the situated cognition paradigm, we introduce Recursive Decoding (RD) -- a new  training and inference paradigm for supervised seq2seq which we apply to gSCAN. Whereas the standard approach is to present an initial world-state, and ask the model to produce an entire action sequence, in RD the model is asked to predict a single action token at a time. Based on this predicted action, the grid-world is then updated to an intermediate state, and given to the model as a fresh input. Instead of formulating the entire action sequence at once, the model is asked to do the ``next best thing'' to move an intermediate world-state closer to the final solution. This allows the model to navigate without having to ``imagine'' what happens as it executes a path, as this computation is effectively offloaded to the environment.

RD dramatically improves performance on the novel direction and length extrapolation gSCAN tasks. In our analyses, we show that these two tasks are related and pose a unique challenge to decoders in Baseline models, which are potentially biased by misleading surface statistics in the target sequences of training examples. RD circumvents these biased surface statistics, and reformulates a complex, out-of-distribution (OOD) sequence generation task into a series of smaller predictions that resemble those seen by the model during training. After presenting our results and analyses, we discuss implications for naturalistic grounded language understanding tasks, and compositional generalization more broadly. We conclude by discussing limitations of our approach and directions for future research.


\section{Methods}

\subsection{gSCAN Overview}

gSCAN is a synthetic dataset for assessing compositional generalization in a grounded language understanding problem ~\cite{ruis2020benchmark}. The model is given two inputs: a command -- specifying a high-level directive in plain english (e.g. ``push the small red square''), and a grid-world -- a 2D matrix representation comprising an agent and various objects with different properties located at different cells. The task of the model is to predict an action sequence that adequately fulfills the input command given a particular grid-world. The target vocabulary consists of the following action tokens: \{\textit{walk}, \textit{turn right/left}, \textit{push}, \textit{pull}, \textit{stay}\}. As this is a seq2seq task, the learning signal consists in how well predicted action sequences correspond to ground-truth action sequences (which we refer to as \textit{target sequences}).

gSCAN comprises a number of curated data splits designed to systematically assess different compositional generalization tasks. Here we are particularly interested in the novel direction and length extrapolation tasks, as both require producing OOD output sequences and have shown poor performance in all previous works. The novel direction task requires generating output sequences that navigate the agent Southwest, whereas in training the target object was never Southwest of the agent's initial position (though it was located in every other relative direction). The length extrapolation task requires generating output sequences that are longer than those seen in training ($n<=15$ in training, $n>15$ in length). A full description of all task splits can be found in~\cite{ruis2020benchmark}.

The dataset includes 367,933 training examples, 3,716 validation examples, 19,282 (in-distribution) test examples, and 88,642 examples in the novel direction task-split. A separate dataset is used to train models on the length task. This contains 180,301 training examples, 1,821 validation examples, 37,784 test examples and 198,588 length-split examples.


\subsection{Baseline Model \& Recursive Decoding}

We use the same baseline model architecture and hyper-parameters reported in \cite{ruis2020benchmark}, except we use a batch size of 512 (instead of 200) for more efficient training. At a high-level, the model consists of an LSTM encoder for the input command and a CNN for encoding the grid-world. The decoder is an attention-based LSTM that attends over both command and grid-world embeddings. The models used on the novel direction and length tasks comprise 391,500 and 535,975 parameters, respectively. Additional parameters for the latter due to a larger CNN kernel size to accommodate the larger grid world in this task. Models were implemented using pyTorch v1.7~\cite{NEURIPS2019_9015}. Code to reproduce all experiments will be made available upon acceptance.

\begin{figure}[h]
\centering
\includegraphics[width=0.6\textwidth]{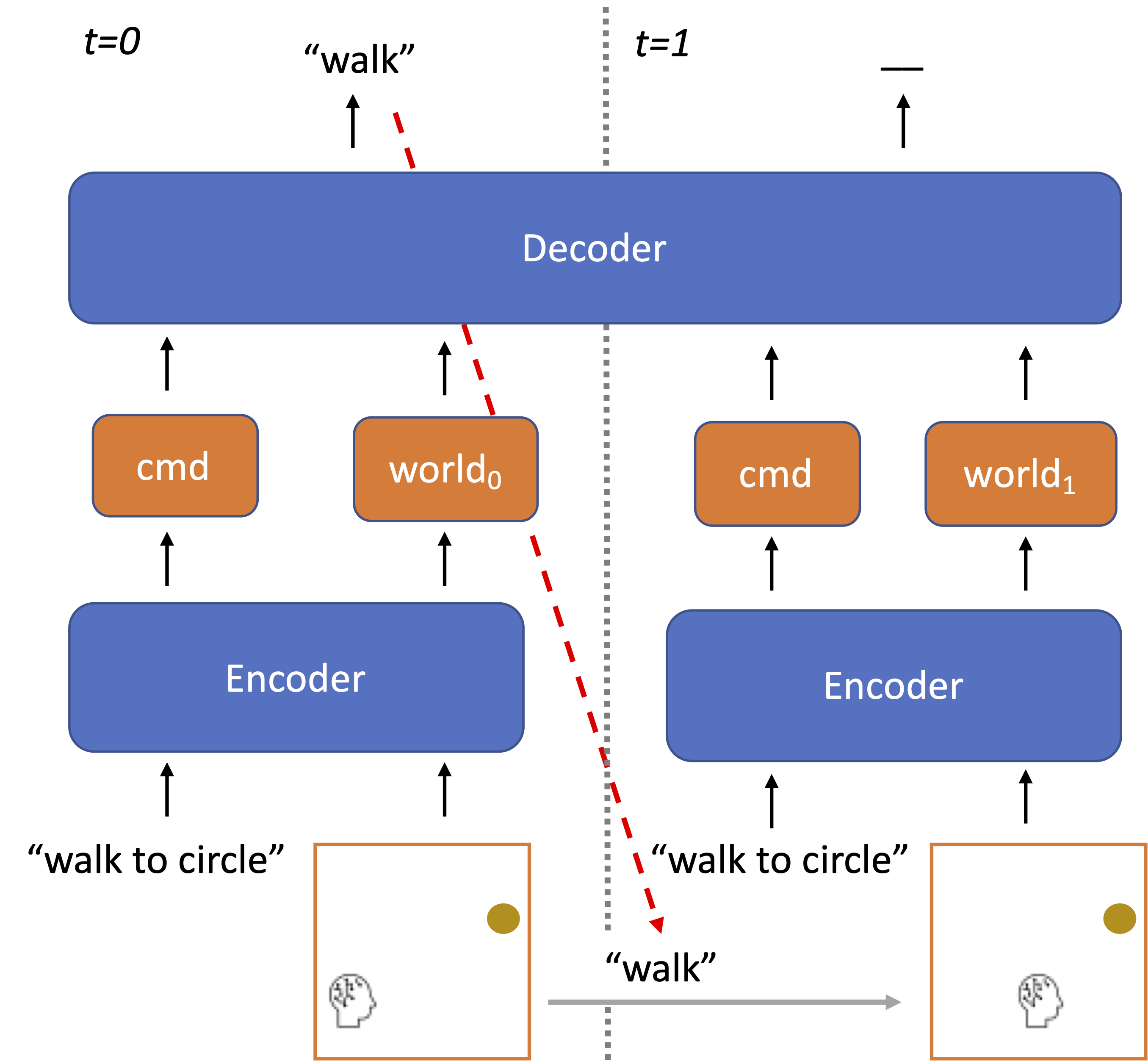}
\caption{Recursive Decoding overview. Rather than predicting a full action sequence all at once based off an initial world-state, the model predicts one action at a time, updates the world state accordingly, and re-encodes the intermediate world-state as if it were a fresh input. This process is repeated until the model predicts EOS.}
\label{fig:recurse-decode}
\end{figure}

The only substantial difference between baseline and RD models in our study is in \textit{how we use} the model during training and inference. Rather than predicting a complete action sequence all at once, we instead train the model to predict one action at a time on the basis of intermediate world-states. After each decoding time-step, we update the world state to reflect the predicted action, and feed this intermediate world-state to the model as a fresh input (see Figure~\ref{fig:recurse-decode}). The model re-encodes this intermediate world-state, and then makes another prediction for the subsequent action. The input command embeddings remain fixed across decoding time, and we allow the hidden state of the decoder LSTM to evolve as it typically would.

As in the baseline approach, we use teacher-forcing during training, such that predictions at each time step are conditioned on ground-truth (as opposed to predicted) action subsequences. In RD, intermediate world states are also updated based on the ground-truth actions during training. Note that we don't perform separate gradient updates after each intermediate prediction; instead gradients are accumulated over all intermediate predictions into a single update. During inference, we update intermediate world states based on predicted actions, and terminate RD once the model predicts an EOS token.

Models were evaluated on a validation set (comprised of in-distribution examples) every 1,000 training batches. For RD models, we implemented an early stopping criteria, such that training stopped if validation loss did not decrease within five validation checks (i.e. 5,000 training batches). We evaluated each experimental condition with three independent training/test runs, and report all results as averages along with standard deviations.\footnote{Baseline models trained on a single GPU in under 24 hours, and RD models trained on a single GPU in under 72 hours. All experiments were executed on a single 16GB P100 GPU. We estimate that altogether 1,200 GPU-hours were required for our experiments.}

\section{Results}

Table~\ref{tbl:results-overview} presents our results across all gSCAN tasks. The RD Random column refers to RD model trained on a dataset where the agent's original orientation is randomized, as described in Section~\ref{sec:novel-direction}. RD dramatically improves performance over state-of-the-art (SotA) on both novel direction and length tasks, but offers no substantial gains in other tasks.\footnote{SotA on Yellow Squares, Relativity, Class inference, and Adverb-to-Verb tasks was achieved by \cite{gao2020systematic}; Red Squares by \cite{heinze2020think}; Novel Direction (prior to RD) and Adverb $(k=1)$ by \cite{kuo2020compositional}. The baseline model presented in \cite{ruis2020benchmark} achieved SotA for Length task (prior to RD).}

\begin{table*}[h]
\caption{Recursive Decoding (RD) performance across gSCAN tasks. Compared to state-of-the-art (SotA), RD offers substantial improvement on Novel Direction and Length tasks.}
\label{tbl:results-overview}
\vskip 0.15in
\begin{center}
\begin{small}
\begin{sc}
\begin{tabular}{lllllr} 
 \toprule
 Task & Baseline & SotA & RD & RD Random \\
 \midrule
 Random & $97.69 \pm 0.22$ & $98.6\pm0.95$ & $99.22\pm0.16$ & $98.39\pm0.17$ \\
 Yellow Squares & $54.96\pm39.39$ & $99.08\pm0.69$ & $82.28\pm11.50$ & $62.19\pm24.08$ \\
 Red Squares & $23.51\pm21.82$ & $81.07\pm10.12$ & $56.29\pm7.42$ & $56.52\pm29.70$ \\
 Novel Direction & $0.00\pm0.00$ & 5.73 & $3.11\pm0.87$ & $\mathbf{43.60\pm6.05}$ \\
 Relativity & $35.02\pm2.35$ & $87.32\pm27.38$ & $57.99\pm7.21$ & $53.89\pm5.39$ \\
 Class inference & $92.52\pm6.75$ & $99.33\pm0.46$ & $98.51\pm0.29$ & $95.74\pm0.75$ \\
 Adverb ($k=1$) & $0.00\pm0.00$ & 11.94 & $0.00\pm0.00$ & $0.00\pm0.00$ \\
 Adverb to verb & $22.70\pm4.59$ & $33.6\pm20.81$ & $21.94\pm0.15$ & $21.95\pm0.03$ \\
 Length & $2.10\pm0.05$ & $2.10\pm0.05$ & $\mathbf{84.42\pm3.24}$ & $79.27\pm0.38$ \\
 \bottomrule
\end{tabular}
\end{sc}
\end{small}
\end{center}
\vskip -0.1in
\end{table*}

An experimental control confirmed that the observed performance gains were due to the procedural innovation of RD, and not a side-effect of greater training variability introduced by the intermediate world states. We created an ``intermediate world'' dataset, which had the same variability in world states that RD models were exposed to in training (see Section~\ref{sec:rd-experimental-control} for details). When trained on this dataset, the baseline model achieved $95\pm0.76$\% on the in-distribution test set, but only $0.29\pm0.23$\% on the novel direction task.

In the next sections we take a deeper dive into the novel direction and length tasks, analyze why and how the baseline model fails, and why Recursive Decoding exhibits such dramatic performance gains.

\subsection{Novel Direction Task}
\label{sec:novel-direction}

\subsubsection{Failure Analysis of Baseline Model}

\begin{figure*}[t]
\centering
\includegraphics[width=\linewidth]{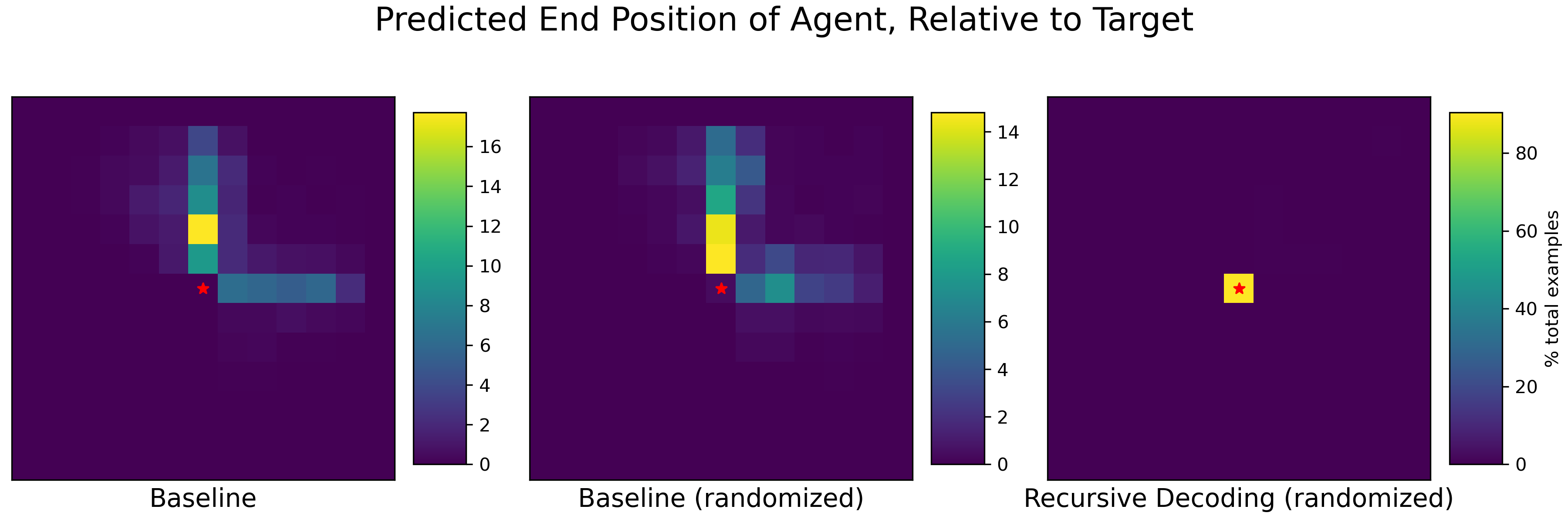}
\caption{Recursive decoding dramatically improves performance on the novel direction task. Heatmaps represent the relative ending position of the agent, as predicted by the model, with respect to the target position. Color indicates percentage of evaluated examples ending in a particular relative position. For simplicity, this analysis was restricted to ``simple walking'' examples, in which the input commands contained no adverbs and no push/pull commands.}
\label{fig:novel-direction-results}
\end{figure*}

The novel direction task is comprised of held-out examples where the target object is Southwest of the agent's position. As stated in \cite{ruis2020benchmark}, despite achieving 0\% exact match on this task, the baseline successfully attends to the correct target cell. Moreover, as depicted in Figure~\ref{fig:novel-direction-results} (leftmost plot), we replicate the finding that in many cases the baseline predicts action sequences that navigate the agent to the correct row or column, but never to the correct target cell. In other words, the model is capable of moving the agent appropriately South \textit{or} West, but never appropriately Southwest.

This could be due to biased surface-statistics in training target sequences. In the original dataset, agents are always initially facing East, so navigating Southwest requires a particular combination of ``walking'' and ``(left/right) turn'' actions that the model never encountered during training. To remedy this, we created another dataset with randomized initial agent orientations. This randomization was designed to disrupt this training set bias because, e.g., the same action sequence can move an agent Northeast or Southwest depending on its starting orientation.

However, when trained and evaluated on the randomized dataset, the model achieved only $0.28\%\pm0.11$ exact match on novel direction examples. Like the original baseline, the agent usually navigated to the correct row or column, but rarely the correct target cell (Figure~\ref{fig:novel-direction-results}, center plot). The slight improvement over the baseline's $0\%\pm0.00$ performance is potentially notable, as it suggests that mitigating surface statistics may have had some effect on generalization for the novel direction task.

\subsubsection{Recursive Decoding performance}

With these results in mind, we were motivated to apply RD to the novel direction task. Intuitively, RD seems well suited for this kind of generalization. The baseline model already demonstrates the ability to navigate sufficiently South or sufficiently West. Once such an intermediate state is reached, the problem resembles an in-distribution training example. RD capitalizes on this by re-encoding intermediate world states as fresh inputs, which should allow the model to simply move in the single remaining direction to reach the target.

We trained and evaluated RD models on both the original and randomized datasets. As indicated in Table~\ref{tbl:results-overview}, the latter yielded significant performance gains, so we focus our discussion around this latter model. RD model achieved an exact match of 43.60\%, a substantial improvement over previous SotA (5.73\% ~\cite{ruis2020benchmark}).

When visualizing the predicted ending position of the agent with respect to the target (Figure~\ref{fig:novel-direction-results}, rightmost plot), it becomes clear that this exact match statistic actually underrepresents model performance, as the model correctly navigates to the target a high percentage of the time. This discrepancy is accounted for by \textit{non-canonical solutions} produced by the RD model. There are many valid action sequences that can correctly satisfy a given input command, even though models are only trained and evaluated against a single, canonical target sequence. 

We evaluated non-canonical solutions on a subset of ``simple walking'' examples from the novel direction task, which had no adverbs or push/pull input commands. The simplicity of these examples allowed us to easily identify non-canonical solutions by determining whether the ending position of the agent matched that of the target object. When we expand our success criteria to include non-canonical solutions, the RD model produces valid solutions for $86.32\pm9.00\%$ of simple walking novel direction examples (37.45\% of which were non-canonical solutions). By contrast, the random-orientation baseline model produced virtually no non-canonical solutions (2/3 models produced 0 non-canonical solutions, and the third produced non-canonical solutions in 0.01\% of simple walking examples).

As described in Appendix~\ref{sec:novel-direction-non-canonical}, some non-canonical solutions were ``error-correcting'' – during decoding, the RD model would predict an incorrect action leading the agent in the wrong direction, but subsequently compensate for this error by re-orienting the agent towards the target (see Figure~\ref{fig:non-canonical-error-correct}). A null model analysis confirmed that these error correcting solutions were more targeted than random walks terminating once the agent happened to arrive at the target. Thus we interpret RD models' ability to produce non-canonical solutions as an adaptive advantage over the baseline.

\subsection{Experimental Control for Recursive Decoding}
\label{sec:rd-experimental-control}

An inevitable side-effect of Recursive Decoding is that models are exposed to a greater variability of inputs during training, because every example comprises a number of intermediate world-states. Thus we tested a control condition to confirm that the observed performance gains were due to the procedural innovation of RD, and not a side-effect of the greater variability of inputs seen during RD training. 

We created an ``intermediate world'' dataset comprising the range of inputs seen by RD models during training. For each example in the original dataset (with randomized starting orientations), we generated all intermediate world-states corresponding to completion of each action in the target sequence. We then packaged up each intermediate world-state with the original input command and the corresponding target sequence, as standalone examples.

Three instances of the Baseline model were trained for 200,000 batches on this intermediate world dataset. The baseline achieved an average exact match of $95.06\pm0.76\%$ on the in-distribution test set, but only $0.29\pm0.23\%$ on the novel direction task, confirming that the reported performance gains were genuinely due to the procedural innovation of Recursive Decoding, and not a side-effect of greater variability of training inputs.

\subsection{Length Extrapolation Task}

Length extrapolation is another generalization task that previous approaches to gSCAN have either ignored, or shown poor performance on (2.10\% SotA ~\cite{kuo2020compositional}). RD yielded substantial performance gains, achieving $84.42\pm3.24\%$ exact match. As with novel direction, there is an intuitive explanation for why RD is so well-suited for length extrapolation. Taken as a whole, an example from the length task is OOD in the sense that a model is required to generate a longer action sequence than it was trained to generate. However, RD decomposes a length example down into a series of individual predictions, which each resemble what the model was exposed to during training.

\subsubsection{Template Analysis of Baseline Failure}

To gain better insight into why and how the baseline model fails, we examined model outputs using a template analysis, in which we reduced predicted and ground-truth action sequences down into high-level template patterns (see Appendix~\ref{sec:template-analysis} for complete details). To do this, we reduce duplicated actions, e.g. ``turn left, turn left" into a single ``turn" action. This allowed us to reduce all action sequences down to a small set of templates consisting of: ``walk push/pull'', ``walk turn walk'' and ``walk turn walk push/pull''. Additionally, we contrasted the counts of tokens in predicted versus ground-truth sequences, and assessed whether predicted sequences navigated agents in the correct direction. Our analysis thus decomposed performance on the length task in terms of whether the predicted action sequence (i) had the proper high-level structure (ii) contained the proper counts of action tokens and (iii) navigated agents in the proper direction.

This analysis revealed that despite its low exact match rate, outputs generated by the baseline model often had the correct high-level structure (on 72\% of examples), and navigated agents in the correct direction (80\%). In most cases, the predicted outputs differed from ground-truth sequences only in that they contained too few action tokens (e.g., the model might predict ``walk walk turn left walk'' instead of ``walk walk walk walk turn left walk''), which reflects its training on shorter target sequences (see Figure~\ref{fig:template-counts-predict-vs-target} for more information).


This type of failure suggests that the baseline model may have been properly representing the underlying structure of length task examples, but failing to perform the internal bookkeeping necessary to track the agent's current position based on its previous outputs. We hypothesize that in training, the baseline decoder was biased to not generate too many consecutive tokens of a given type (e.g., don't ``turn right'' after predicting four ``walk'' tokens), resulting in an inability to generalize to longer sequences in the length task. By contrast, RD models did not have to keep track of token-count as they decoded, because agent position was explicitly reflected in intermediate world states. Thus, RD models may have avoided internalizing the biased token-count surface statistics in training examples.

\subsubsection{Hidden-State Analysis}
\label{sec:hidden-state-analysis}

A hidden state analysis of the LSTM decoders in baseline and RD models was performed to test the above hypothesis. The analysis was driven by two questions: (1) Is time-step explicitly encoded in decoder hidden states? and (2) How (if at all) do hidden states corresponding to early time-steps differ from late (OOD) time-steps? We expect that baseline models will explicitly encode time-step and that representations for early and late time-steps will be distinct. In contrast, we hypothesize that the decoder trained using RD will not track time-step and thus exhibit no meaningful difference in hidden states of early versus late time-steps.

\begin{figure}[h]
\centering
\includegraphics[width=.7\linewidth]{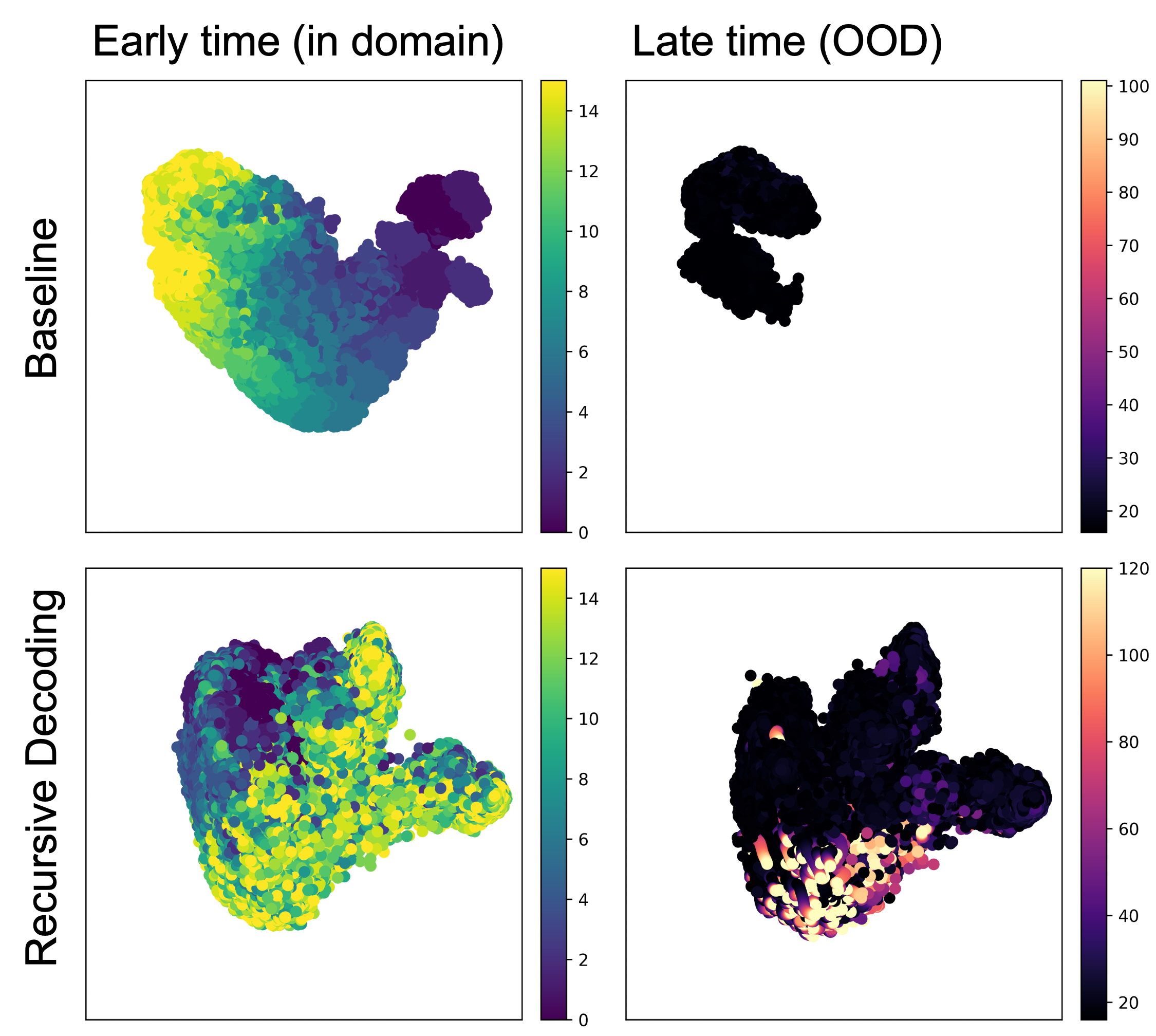}
\caption{Hidden states of length extrapolation examples. The top two principal components explain 33.2\% of the variance for Baseline model (top) and 35.9\% for Recursive Decoding (bottom). Decoder time-step is indicated by color scale. Early hidden states (in-distribution, $t<=15$) are shown on the left, and late states (out-of-distribution, $t>15$) are shown on the right. PCA was performed over all early and late hidden states for each model.}
\label{fig:hidden-state}
\end{figure}

Principal Component Analysis (PCA) was performed on the set of 100-dimensional hidden states at each decoding time-step for all length-split examples. Dimensional reductions are shown in Figure~\ref{fig:hidden-state}. Top and lower plots depict states for baseline and RD decoders respectively (PCA was separately performed on hidden states of each model).  We plot the first two principal components, which accounted for 33.2\% total variance in baseline hidden states, and 35.9\% of RD states. Decoding time is indicated by color. Left plots display hidden states corresponding to early (in domain) time-steps $(t<=15)$ and right plots display hidden states of late (OOD) time-steps $(t>15)$. Early and late time-steps are plotted separately for visual clarity, but they share the same components (within each model) because PCA was conducted on the joint set of states from all time-steps.

These plots reveal two striking differences in the behavior of baseline versus RD decoders. First, the baseline decoder appears to encode time-step, as hidden states corresponding to the same time-step form tight clusters. In the RD model, although it is roughly correlated, the time-steps are widely dispersed throughout the latent space.

This observation was quantitatively verified with a probe analysis. Multi-layer perceptrons (MLP) were trained to predict time-step from the hidden states of each model. MLPs consisted of two linear layers; the first projected the original hidden states down into a 50-dimensional embedding, and the second projected them down to a single scalar which was regressed against time-step using Mean Squared Error. Models were separately trained on baseline and RD data for 50 epochs, with a batch size of 512. The best-performing models predicted time-step of held-out examples to within $\pm1$ of ground truth 99.54\% of the time for baseline model, but only 22.43\% of the time for RD model.

The second important difference is that in the baseline decoder, OOD hidden states (Figure~\ref{fig:hidden-state}, top-right) cluster around a small region of latent space that is distinct from the majority of in-distribution hidden states (top-left plot). Thus, as decoding time-step increases past training distribution, the hidden-state of baseline decoders is pushed into a restricted OOD region of latent space from which it is unable to escape. This is consistent with recent demonstrations that seq2seq decoders are pulled into limited ``attractor'' regions of state space when forced to generate past the length of training sequences, precluding their ability to generate meaningful continuations \cite{newman2020eos}. Conversely, in the RD decoder, hidden-states of late OOD time-steps (bottom-right) overlap with the same region of latent space as in-distribution states (bottom-left).

Taken together, these results provide support for our initial hypothesis. Baseline models explicitly encode time-step as they decode, resulting in degenerate dynamics once decoding time extends beyond the training distribution. Conversely, RD decoders have less explicit encoding of time-step, allowing late time-steps to be treated similarly to early time-steps. Baseline models are required to track their previous outputs as they decode, because they need this information to infer the current position of an agent relative to the target object. RD appears to offload this computation to the external environment; RD models don't have to infer current position of an agent because this information is explicitly encoded in intermediate world-states.

\section{Related Work}

Since its release, there has been limited work on gSCAN. Previous approaches have focused on changes to model architecture that support more refined representations of input commands and world state. For instance, by modeling objects in the grid-world as a network and leveraging message-passing to refine grid-world embeddings~\cite{gao2020systematic}. Another approach employed modular networks, composed in different arrangements as a function of the input command \cite{kuo2020compositional}. Lastly, \cite{heinze2020think} decomposed gSCAN down into two stages: (i) identifying the target object and (ii) navigating to said target. As reflected in the SotA column of Table~\ref{tbl:results-overview}, these approaches have increased performance on gSCAN tasks requiring compositional encoding of novel inputs, but offer no substantial improvement on the novel direction and length tasks, which require decoders to generate OOD action sequences. This underscores the point that compositionality comes in different forms, which are amenable to different modeling approaches \cite{hupkes2020compositionality}.

RD is particularly well-suited to ``decode-side'' generalization, because it recursively decomposes a complex, OOD sequence generation task down into a series of smaller predictions that each appear in-distribution from the perspective of the model. In this spirit, RD is closely related to the iterative decoding methodology recently introduced by \cite{ruiz2021iterative}, who trained transformer models to compositionally ``unroll'' a complex seq2seq problem into a series of sub-steps, which are iteratively solved one at a time. The primary difference is that RD is \textit{recursive}, because the output sequence at any time-step is composed of the current predicted token appended to the history of predicted tokens from previous RD time-steps. RD also resembles Imitation Learning, within the Reinforcement Learning paradigm, in that both approaches involve predicting actions in response to evolving environmental feedback, and learning from an ``expert'' canonical target sequence \cite{abbeel2004apprenticeship}. Here we show that the benefits of such environmental feedback can be enjoyed in the context of supervised seq2seq. 

More broadly, RD parallels current trends in open-domain multi-hop question answering, where the state of art has moved to iteratively querying batches of source documents from a corpus \cite{xiong2020answering, qi2020retrieve}. While a different domain from grounded language understanding, this reflects an emerging recognition that a single forward-pass is often insufficient for complex, multi-stage reasoning or planning tasks.

\section{Discussion}

This paper presented Recursive Decoding (RD), a novel approach to seq2seq. RD was applied to gSCAN, a grounded language compositional generalization benchmark, and dramatically improved performance on two previously neglected tasks: novel direction and length extrapolation. These generalization tasks pose particular challenges, as they require generating OOD output sequences, composed of novel combinations of in-distribution output subsequences, from in-distribution input sequences. Whereas other approaches have focused on modifying the training objective \cite{lake2019compositional, baan2019realization}, model architecture \cite{kuo2020compositional, gao2020systematic}, or data augmentation \cite{andreas2019good, akyurek2020learning}, we use RD to reformulate and decompose a complex sequence generation task into a series of smaller predictions that appear in-distribution from the perspective of the model. 

We draw inspiration from situated cognition in the context of grounded language understanding, and show that intermediate states in the external world can be leveraged to help incrementally guide models to a correct solution. In so doing, much of the bookkeeping that traditional models have to do (i.e., of ``imagining'' an agent's current position based on its history of predicted actions) is offloaded onto the external environment. As observed in Section~\ref{sec:hidden-state-analysis}, this has the additional advantage of eliminating reliance on biased surface cues, such as token count in the length task, that otherwise deter models from generating OOD output sequences. These benefits reaffirm recent demonstrations that disentangling seq2seq models' underlying task representations from the bookkeeeping necessary to track their progress in relation to a final solution makes them more amenable to compositional generalization \cite{ma2019self, zheng2021disentangled}.

Unlike gSCAN, in which the model has access to a birds-eye-view map of an idealized environment, most naturalistic grounded language tasks entail egocentric navigation, where agents get local visual feedback based on their position in an external environment \cite{anderson2018vision, shridhar2020alfred}. In these cases, the benefits of environmental offloading are baked into the problem structure and don't need to be explicitly engineered. While not a call to action, our results suggest this may be advantageous for future modeling approaches.

More generally, we identify the novel direction and length extrapolation gSCAN tasks as a particular kind of compositional generalization task that requires generating OOD output sequences. This ``decode-side'' flavor of compositionality is germane to many seq2seq tasks, including visual question answering/automated captioning, code generation, and length extrapolation in seeded text generation \cite{goyal2017making, chen2021evaluating, newman2020eos}. As we show, the difficulty presented by such tasks can sometimes be ameliorated by reformulating the problem into a series of incremental sub-sequence generations which resemble what models see during training. As formulated in this paper, RD is specific to gSCAN, but in the future this general approach could be adapted to other tasks and datasets. The key challenge to such an adaptation would be finding (or constructing) an analogue to the gSCAN grid world; a supplemental object whose state can be incrementally updated on the basis of predicted outputs.

\section{Future Directions and Limitations}

The most straightforward disadvantage of RD is that it is slow; a single example requires $n$ forward passes through the model, where $n$ is the length of the output sequence. Future work might investigate how best to retain the benefits of RD while increasing computational efficiency.

A second limitation of our results is that we presume the ability to update the world state based on model actions. While this might be trivial for gSCAN because of its limited nature, more complex tasks might not support oracle predictions of world state. In such a scenario, the model would need to be updated either to generate its own world state predictions or to operate over uncertain or noisy states.

\bibliographystyle{plain}
\bibliography{main}

\begin{thebibliography}{10}

\bibitem{abbeel2004apprenticeship}
Pieter Abbeel and Andrew~Y Ng.
\newblock Apprenticeship learning via inverse reinforcement learning.
\newblock In {\em Proceedings of the twenty-first international conference on
  Machine learning}, page~1, 2004.

\bibitem{akyurek2020learning}
Ekin Aky{\"u}rek, Afra~Feyza Aky{\"u}rek, and Jacob Andreas.
\newblock Learning to recombine and resample data for compositional
  generalization.
\newblock {\em arXiv preprint arXiv:2010.03706}, 2020.

\bibitem{anderson2018vision}
Peter Anderson, Qi~Wu, Damien Teney, Jake Bruce, Mark Johnson, Niko
  S{\"u}nderhauf, Ian Reid, Stephen Gould, and Anton Van Den~Hengel.
\newblock Vision-and-language navigation: Interpreting visually-grounded
  navigation instructions in real environments.
\newblock In {\em Proceedings of the IEEE Conference on Computer Vision and
  Pattern Recognition}, pages 3674--3683, 2018.

\bibitem{andreas2019good}
Jacob Andreas.
\newblock Good-enough compositional data augmentation.
\newblock {\em arXiv preprint arXiv:1904.09545}, 2019.

\bibitem{baan2019realization}
Joris Baan, Jana Leible, Mitja Nikolaus, David Rau, Dennis Ulmer, Tim
  Baumg{\"a}rtner, Dieuwke Hupkes, and Elia Bruni.
\newblock On the realization of compositionality in neural networks.
\newblock {\em arXiv preprint arXiv:1906.01634}, 2019.

\bibitem{calvo2008handbook}
Paco Calvo and Toni Gomila.
\newblock {\em Handbook of cognitive science: An embodied approach}.
\newblock Elsevier, 2008.

\bibitem{chen2021evaluating}
Mark Chen, Jerry Tworek, Heewoo Jun, Qiming Yuan, Henrique Ponde de~Oliveira
  Pinto, Jared Kaplan, Harri Edwards, Yuri Burda, Nicholas Joseph, Greg
  Brockman, et~al.
\newblock Evaluating large language models trained on code.
\newblock {\em arXiv preprint arXiv:2107.03374}, 2021.

\bibitem{chiel1997brain}
Hillel~J Chiel and Randall~D Beer.
\newblock The brain has a body: adaptive behavior emerges from interactions of
  nervous system, body and environment.
\newblock {\em Trends in neurosciences}, 20(12):553--557, 1997.

\bibitem{chomsky2009syntactic}
Noam Chomsky.
\newblock {\em Syntactic structures}.
\newblock De Gruyter Mouton, 2009.

\bibitem{clark1998extended}
Andy Clark and David Chalmers.
\newblock The extended mind.
\newblock {\em analysis}, 58(1):7--19, 1998.

\bibitem{dankers2021paradox}
Verna Dankers, Elia Bruni, and Dieuwke Hupkes.
\newblock The paradox of the compositionality of natural language: a neural
  machine translation case study.
\newblock {\em arXiv preprint arXiv:2108.05885}, 2021.

\bibitem{gao2020systematic}
Tong Gao, Qi~Huang, and Raymond~J Mooney.
\newblock Systematic generalization on gscan with language conditioned
  embedding.
\newblock {\em arXiv preprint arXiv:2009.05552}, 2020.

\bibitem{goyal2017making}
Yash Goyal, Tejas Khot, Douglas Summers-Stay, Dhruv Batra, and Devi Parikh.
\newblock Making the v in vqa matter: Elevating the role of image understanding
  in visual question answering.
\newblock In {\em Proceedings of the IEEE Conference on Computer Vision and
  Pattern Recognition}, pages 6904--6913, 2017.

\bibitem{heinze2020think}
Christina Heinze-Deml and Diane Bouchacourt.
\newblock Think before you act: A simple baseline for compositional
  generalization.
\newblock {\em arXiv preprint arXiv:2009.13962}, 2020.

\bibitem{hupkes2020compositionality}
Dieuwke Hupkes, Verna Dankers, Mathijs Mul, and Elia Bruni.
\newblock Compositionality decomposed: how do neural networks generalise?
\newblock {\em Journal of Artificial Intelligence Research}, 67:757--795, 2020.

\bibitem{kamp1995prototype}
Hans Kamp and Barbara Partee.
\newblock Prototype theory and compositionality.
\newblock {\em Cognition}, 57(2):129--191, 1995.

\bibitem{kim2020cogs}
Najoung Kim and Tal Linzen.
\newblock Cogs: A compositional generalization challenge based on semantic
  interpretation.
\newblock {\em arXiv preprint arXiv:2010.05465}, 2020.

\bibitem{kuo2020compositional}
Yen-Ling Kuo, Boris Katz, and Andrei Barbu.
\newblock Compositional networks enable systematic generalization for grounded
  language understanding.
\newblock {\em arXiv preprint arXiv:2008.02742}, 2020.

\bibitem{lake2018generalization}
Brenden Lake and Marco Baroni.
\newblock Generalization without systematicity: On the compositional skills of
  sequence-to-sequence recurrent networks.
\newblock In {\em International conference on machine learning}, pages
  2873--2882. PMLR, 2018.

\bibitem{lake2014towards}
Brenden~M Lake.
\newblock {\em Towards more human-like concept learning in machines:
  Compositionality, causality, and learning-to-learn}.
\newblock PhD thesis, Massachusetts Institute of Technology, 2014.

\bibitem{lake2019compositional}
Brenden~M Lake.
\newblock Compositional generalization through meta sequence-to-sequence
  learning.
\newblock {\em arXiv preprint arXiv:1906.05381}, 2019.

\bibitem{lake2019human}
Brenden~M Lake, Tal Linzen, and Marco Baroni.
\newblock Human few-shot learning of compositional instructions.
\newblock {\em arXiv preprint arXiv:1901.04587}, 2019.

\bibitem{ma2019self}
Chih-Yao Ma, Jiasen Lu, Zuxuan Wu, Ghassan AlRegib, Zsolt Kira, Richard Socher,
  and Caiming Xiong.
\newblock Self-monitoring navigation agent via auxiliary progress estimation.
\newblock {\em arXiv preprint arXiv:1901.03035}, 2019.

\bibitem{marghetis2016mastering}
Tyler Marghetis, David Landy, and Robert~L Goldstone.
\newblock Mastering algebra retrains the visual system to perceive hierarchical
  structure in equations.
\newblock {\em Cognitive research: principles and implications}, 1(1):1--10,
  2016.

\bibitem{newman2020eos}
Benjamin Newman, John Hewitt, Percy Liang, and Christopher~D Manning.
\newblock The eos decision and length extrapolation.
\newblock {\em arXiv preprint arXiv:2010.07174}, 2020.

\bibitem{NEURIPS2019_9015}
Adam Paszke, Sam Gross, Francisco Massa, Adam Lerer, James Bradbury, Gregory
  Chanan, Trevor Killeen, Zeming Lin, Natalia Gimelshein, Luca Antiga, Alban
  Desmaison, Andreas Kopf, Edward Yang, Zachary DeVito, Martin Raison, Alykhan
  Tejani, Sasank Chilamkurthy, Benoit Steiner, Lu~Fang, Junjie Bai, and Soumith
  Chintala.
\newblock Pytorch: An imperative style, high-performance deep learning library.
\newblock In H.~Wallach, H.~Larochelle, A.~Beygelzimer, F.~d\textquotesingle
  Alch\'{e}-Buc, E.~Fox, and R.~Garnett, editors, {\em Advances in Neural
  Information Processing Systems 32}, pages 8024--8035. Curran Associates,
  Inc., 2019.

\bibitem{qi2020retrieve}
Peng Qi, Haejun Lee, Oghenetegiri Sido, and Christopher~D Manning.
\newblock Retrieve, rerank, read, then iterate: Answering open-domain questions
  of arbitrary complexity from text.
\newblock {\em arXiv e-prints}, pages arXiv--2010, 2020.

\bibitem{ruis2020benchmark}
Laura Ruis, Jacob Andreas, Marco Baroni, Diane Bouchacourt, and Brenden~M Lake.
\newblock A benchmark for systematic generalization in grounded language
  understanding.
\newblock {\em arXiv preprint arXiv:2003.05161}, 2020.

\bibitem{ruiz2021iterative}
Luana Ruiz, Joshua Ainslie, and Santiago Onta{\~n}{\'o}n.
\newblock Iterative decoding for compositional generalization in transformers.
\newblock {\em arXiv preprint arXiv:2110.04169}, 2021.

\bibitem{shridhar2020alfred}
Mohit Shridhar, Jesse Thomason, Daniel Gordon, Yonatan Bisk, Winson Han,
  Roozbeh Mottaghi, Luke Zettlemoyer, and Dieter Fox.
\newblock Alfred: A benchmark for interpreting grounded instructions for
  everyday tasks.
\newblock In {\em Proceedings of the IEEE/CVF conference on computer vision and
  pattern recognition}, pages 10740--10749, 2020.

\bibitem{smith2005development}
Linda Smith and Michael Gasser.
\newblock The development of embodied cognition: Six lessons from babies.
\newblock {\em Artificial life}, 11(1-2):13--29, 2005.

\bibitem{xiong2020answering}
Wenhan Xiong, Xiang~Lorraine Li, Srini Iyer, Jingfei Du, Patrick Lewis,
  William~Yang Wang, Yashar Mehdad, Wen-tau Yih, Sebastian Riedel, Douwe Kiela,
  et~al.
\newblock Answering complex open-domain questions with multi-hop dense
  retrieval.
\newblock {\em arXiv preprint arXiv:2009.12756}, 2020.

\bibitem{zheng2021disentangled}
Hao Zheng and Mirella Lapata.
\newblock Disentangled sequence to sequence learning for compositional
  generalization.
\newblock {\em arXiv preprint arXiv:2110.04655}, 2021.

\end{thebibliography}

\clearpage

\appendix

\counterwithin{figure}{section}





\section{Non-Canonical Solutions to Novel Direction task}
\label{sec:novel-direction-non-canonical}

\begin{figure}[h]
\centering
\includegraphics[width=.5\linewidth]{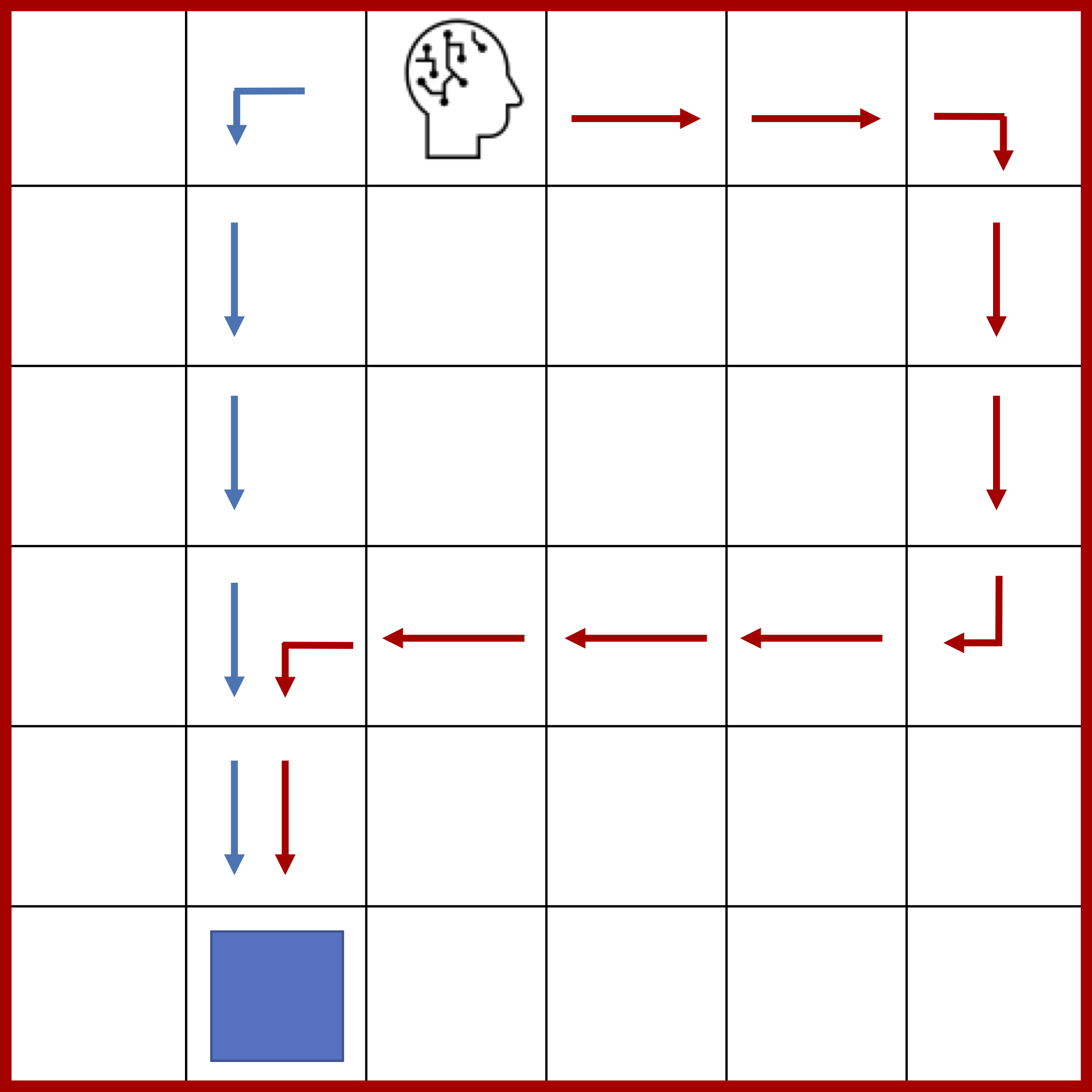}
\caption{Error-correcting solution generated by the Recursive Decoding model. Predicted action sequence is shown in red, and the canonical path is shown in blue. The model initially misleads the agent in the wrong direction, but subsequently compensates for this error by re-orienting the agent back towards the target object.}
\label{fig:non-canonical-error-correct}
\end{figure}

The difference between canonical and non-canonical solutions was often arbitrary (e.g., two right turns instead of two left turns). However, some non-canonical solutions were ``error correcting'' – during decoding, RD models would predict an incorrect action that would lead the agent in the wrong direction, but subsequently compensate for this error by reversing the direction of the agent back towards the target object. An example of such an error-correcting solution is shown in Figure~\ref{fig:non-canonical-error-correct}. Red arrows depict the non-canonical solution produced by an RD model, and blue arrows depict the canonical action sequence.

We constructed a simple null model to ensure that non-canonical solutions to novel direction examples produced by the RD model were more targeted than trivial random walks terminating whenever agents coincided with the target. The null model randomly predicted ``walk'', ``turn left'' or ``turn right'' at each time step with uniform probability, and action sequences terminated whenever the agent reached the target cell. This was evaluated on all ``simple walking'' examples in the novel direction data split.

We compared the lengths of RD-predicted non-canonical sequences to random walk sequences (generated by the null model), with the rationale that RD sequences should be shorter if the RD model was genuinely making sense of these examples. Indeed, we found that RD-predicted non-canonical solutions were significantly shorter than those produced by the null model (paired t(3238) = -55.63, $p<0.001$). Compared to canonical target sequences, RD-predicted non-canonical sequences were within 5 tokens of target length on average (25\% were within one token and 50\% were within 4 tokens), whereas random walk sequences were 250 tokens longer than target sequences on average (25\% were within 79 and 50\% were within 176).

\section{Template Analysis for Length Task}
\label{sec:template-analysis}

We performed a ``template analysis'' to better understand how and why the baseline model failed on the length task. In this analysis, predicted and target action sequences were reduced down to high-level template patterns. For example, the action sequence ``walk walk walk turn right walk walk pull pull pull'' would be reduced to ``walk turn walk pull''. This allowed us to go beyond exact match accuracy, and evaluate whether the predicted action sequence had the proper high-level structure.

We also recorded the count of consecutive actions of a given type in a separate data structure (e.g. ``walk walk walk turn turn'' would be converted to $[3\times walk, 2\times turn]$). This allowed us to assess how the counts of predicted actions compared to the ground truth. Lastly, we assessed whether predicted action sequences navigated agents in the proper direction (i.e., towards the target object). Our analysis thus decomposed performance on the length task in terms of whether the predicted action sequence (1) had the proper high-level structure (2) contained the proper counts of action tokens and (3) navigated agents in the proper direction.

Steps for reducing an action sequence down into a template are as follows:
\begin{enumerate}[nolistsep]
    \item Strip all initial turn actions.
    \item Substitute all remaining ``turn left'' and ``turn right'' cmds with ``turn'', and all ``push'' and ``pull'' commands to ``push/pull''.
    \item Condense all consecutive action tokens of same type to a single token (e.g. ``walk walk walk turn turn'' $\rightarrow [walk, turn]$)
    \item Record the number of consecutive action types in a separate data structure (e.g. ``walk walk walk turn turn'' $\rightarrow [3\times walk, 2\times turn]$).
\end{enumerate}

\begin{figure}[h]
\centering
\centering
\includegraphics[width=\linewidth]{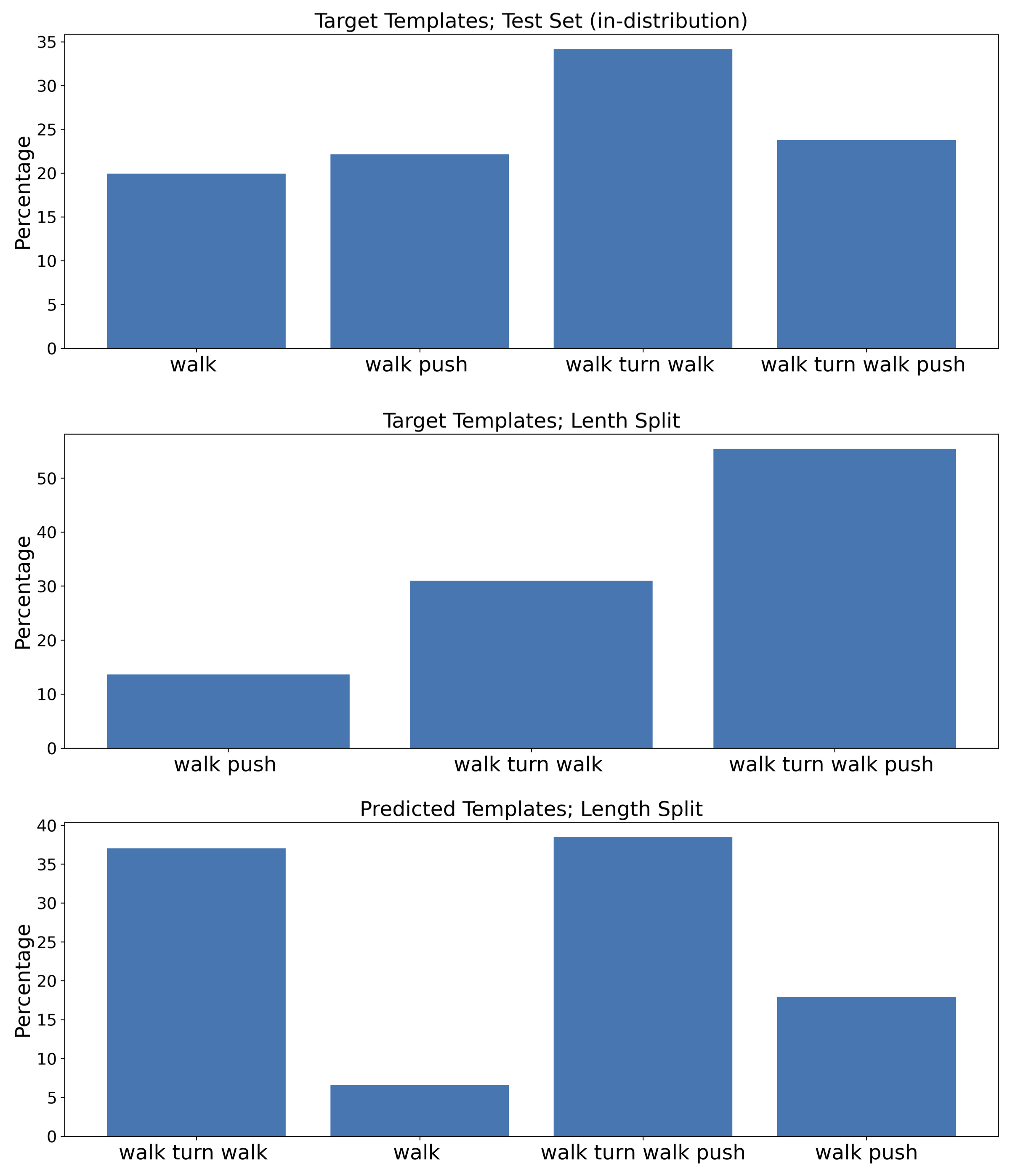}
\caption{Distributions of action sequence templates for in-distribution targets (top), length-split targets (middle) and length-split predictions (bottom). All model predictions correspond to well-formed templates, and the distribution of predicted templates is a sort of comprise between the ground-truth template distributions of length and in-distribution data splits.}
\label{fig:template-distributions}
\end{figure}

Our analysis revealed that all output sequences within the length training/test sets can be categorized into a small number of templates. Figure~\ref{fig:template-distributions} displays the distributions of target sequence (i.e., ground-truth) templates in the training set (top left) and length-split (bottom left), as well as predicted templates for the length-split (right). This illustrates that target sequences in training versus length splits are drawn from different template distributions, and baseline predictions are a sort of compromise between the two.

\begin{figure}[h]
\centering
\includegraphics[width=.6\linewidth]{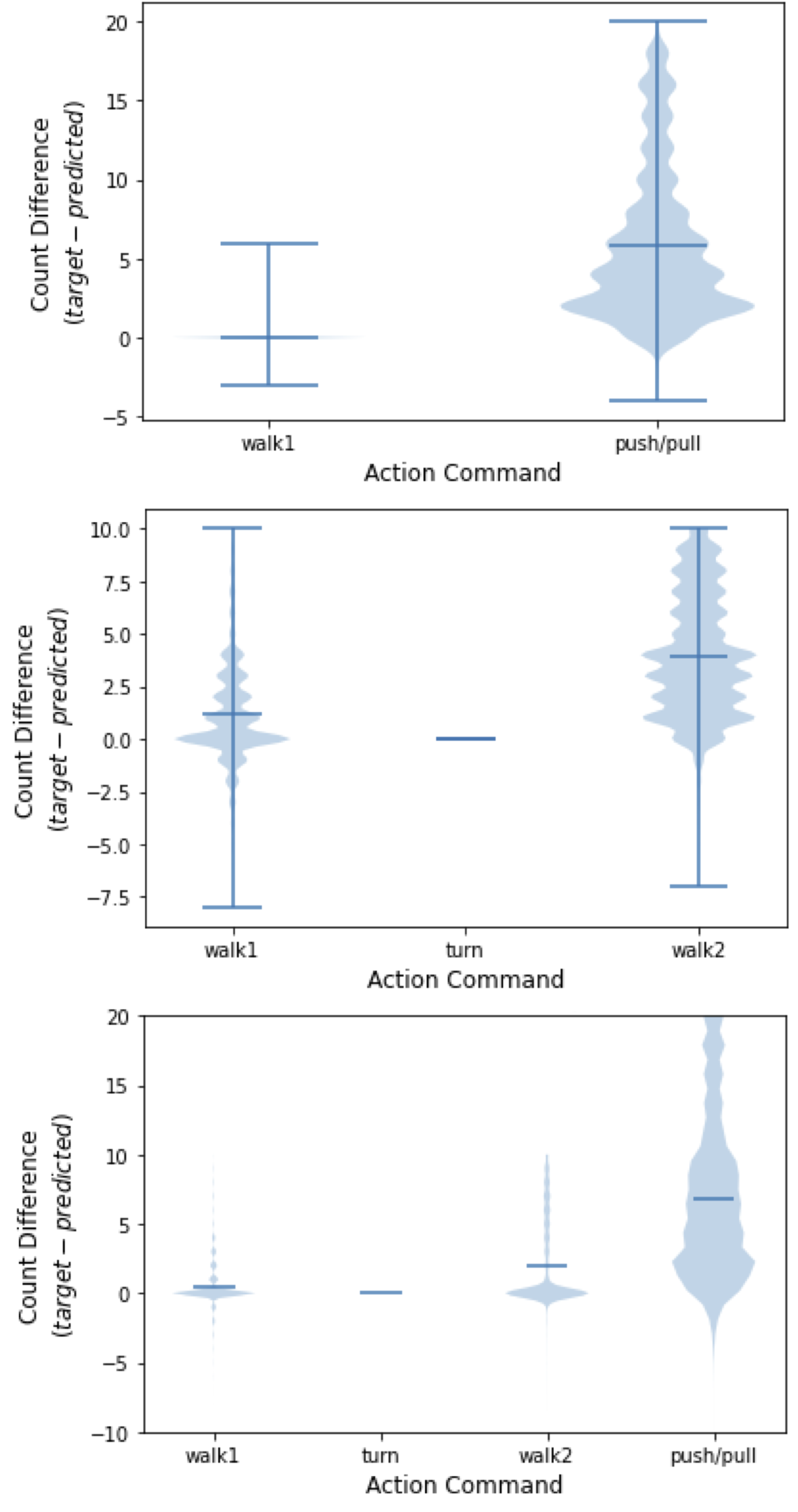}
\caption{Baseline model predicts too few tokens. Token count distributions show the difference in counts of predicted versus ground truth action tokens, by template-index. Positive values indicate higher action-counts in target sequence relative to predicted sequences. Horizontal lines depict means. Distributions are plotted separately for each template (i.e., top plot is ``walk push/pull'' template, middle is ``walk turn walk'', bottom is ``walk turn walk push/pull''). Only examples where the model predicted the correct template are included.}
\label{fig:template-counts-predict-vs-target}
\end{figure}

Overall, the baseline model predicted the correct template on 72\% of all length-split examples, and navigated to the proper direction 80\% of the time (it produced the proper template \textit{and} navigated in the proper direction on 65\% of the examples). Thus, despite its poor performance in terms of exact match, the model often produces well-formed outputs in the Length task, that match the high-level structure of target solutions.

Given this disparity between template and exact match, we analyzed model failure in terms of action token counts in predicted versus target sequences. We examined all examples where the model predicted the correct template, grouped them by template-type, and compared the relative counts in target versus predicted sequences at each template index. Distributions of these relative counts are plotted in Figure~\ref{fig:template-counts-predict-vs-target}, where positive values indicate that the model predicted \textit{too few} actions (middle horizontal line denotes average).

Figure~\ref{fig:template-counts-predict-vs-target} illustrates a few important points. First, models typically produce too few $walk$ and $push/pull$ actions, but the correct number of $turn$ actions. This makes sense because turns only ever appear at most two times consecutively in target sequences, to re-orient an agent 180 degrees. Second, although in aggregate the model exhibits a bias towards producing too few actions, there are substantial modes around zero, indicating that in many examples the model does indeed predict the correct (or close to the correct) number of actions.

\end{document}